\documentclass[twoside,11pt, preprint]{article}

\usepackage{blindtext}

\usepackage[abbrvbib, preprint]{dmlr2e}

\usepackage{lastpage}

\usepackage[T1]{fontenc}
\usepackage{array}
\usepackage{booktabs}
\usepackage{graphicx}
\usepackage{caption}
\usepackage{subcaption}
\usepackage{xcolor}
\usepackage{amssymb}
\usepackage{pifont}
\newcommand{\cmark}{\ding{51}}%
\newcommand{\xmark}{\ding{55}}%
\usepackage{subcaption}
\usepackage{caption}

\usepackage{color, colortbl}
\definecolor{Gray}{gray}{0.9}

\definecolor{verylightgray}{rgb}{0.95, 0.95, 0.95}
\definecolor{verylightgreen}{rgb}{0.9, 1, 0.9}
\definecolor{verylightred}{rgb}{1, 0.9, 0.9}

\usepackage{arydshln}

\def\openreview{}

\begin{document}

\title{Towards Grounded Visual Spatial Reasoning in Multi-Modal Vision Language
Models}

\author{\name Navid Rajabi \email nrajabi@gmu.edu \\
       \addr Department of Computer Science\\
       George Mason University\\
       Fairfax, VA 22030, USA
       \AND
       \name Jana Ko{\v{s}}eck{\'a} \email kosecka@gmu.edu \\
       \addr Department of Computer Science\\
       George Mason University\\
       Fairfax, VA 22030, USA}

\editor{My editor}

\maketitle

\begin{abstract}
Large vision-and-language models (VLMs) trained to match images with text on large-scale datasets of image-text pairs have shown impressive generalization ability on several vision and language tasks. Several recent works, however, showed that these models lack fine-grained understanding, such as the ability to count and recognize verbs, attributes, or relationships. The focus of this work is to study the understanding of spatial relations. This has been tackled previously using image-text matching (e.g., Visual Spatial Reasoning benchmark) or visual question answering (e.g., GQA or VQAv2), both showing poor performance and a large gap compared to human performance. In this work, we show qualitatively (using explainability tools) and quantitatively (using object detectors) that the poor object localization "grounding" ability of the models is a contributing factor to the poor image-text matching performance. We propose an alternative fine-grained, compositional approach for recognizing and ranking spatial clauses that combines the evidence from grounding noun phrases corresponding to objects and their locations to compute the final rank of the spatial clause. We demonstrate the approach on representative VLMs (\textit{such as} LXMERT, GPV, and MDETR) and compare and highlight their abilities to reason about spatial relationships. 
\end{abstract}

\begin{keywords}
Visual Spatial Reasoning, Vision-and-Language Models (VLMs), Multi-Modal Transformers, Fine-grained Grounding, Compositional Understanding.
\end{keywords}

\section{Introduction}


Visual reasoning is among the general goals of vision-and-language models (VLMs). Recent advances in multi-modal vision-language models were enabled by self-supervised pre-training objectives on web-scale datasets of image-text pairs. 
The evaluation methodologies usually resort to measuring performance on downstream tasks 
that include visual question answering (VQA), referring expression comprehension, image-to-text/text-to-image retrieval, image-text matching, or generative/auto-regressive tasks, like image captioning. 
%
Multiple recent studies have shown that vision-language models lack fine-grain understanding of verbs~\cite{svo}, spatial relationship~\cite{vsr, kamath2023s}, word order~\cite{winoground}, and lack general visio-linguistic compositionality~\cite{stanford-bag-of-words} that is critical for compositional reasoning and generalization. 
%
\begin{figure}[!h]
    \centering
    \includegraphics[scale=0.33]{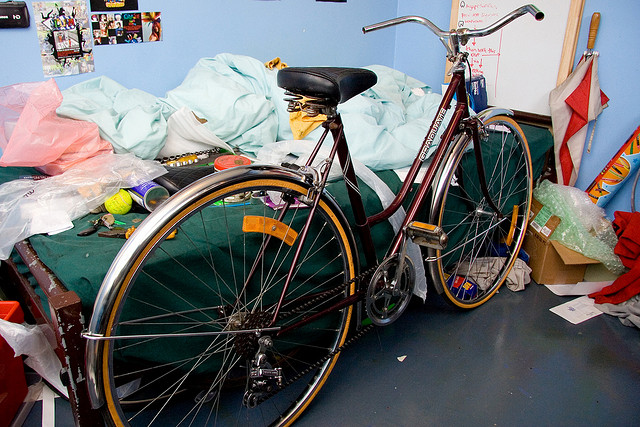}
    \caption{Although the ground-truth caption for this image is: "The \colorbox{green}{\textbf{bed}} is \colorbox{orange}{\textbf{next to}} the \colorbox{yellow}{\textbf{bicycle}}", multiple different \textit{spatial clauses} from the language domain can be inferred from the visual domain (image) as to fill out the spatial clauses correctly, like \textit{behind}, \textit{left of}, \textit{near}, \textit{close to}, \textit{touching}, etc. This type of intrinsic ambiguity from the language side makes formulating the spatial reasoning task more challenging.}
    \label{fig:fig1}
\end{figure}
In this work, we focus on visual spatial reasoning task considered in Visual Spatial Reasoning (VSR) benchmark~\cite{vsr} and show qualitatively, using explainability tools~\cite{explainability}, and quantitatively, using the performance of object detectors, that a significant weakness of multi-modal models is the ability of grounding subjects and objects of the spatial relationships.  
We then propose an alternative compositional ranking-based approach for spatial clause understanding built on the top of General Purpose Vision (GPV) multi-modal model~\cite{gpv} pre-trained with object-level bounding box supervision and classification, visual question answering, and captioning.  
Our contributions can be summarized as follows:\\
\noindent {\bf i)} We analyze the top-performing LXMERT model~\cite{tan2019lxmert}) on VSR benchmark~\cite{vsr} and identify problems with noun grounding and the image-text matching (ITM) evaluation methodology.\\
\noindent {\bf ii)} We decompose spatial clauses into simpler primitives by grounding and localizing {\em subject} and {\em object} using an encoder-decoder GPV model~\cite{gpv} and train a separate, lightweight model for predicting their {\em spatial relation}. This yields a more traditional structured approach for combining the evidence about objects and spatial relations and ranking the spatial clauses.\\
\noindent {\bf iii)} The approach is evaluated on a subset of the VSR benchmark, demonstrating the effectiveness of our strategy for spatial clause understanding and the suitability of our ranking-based approach that avoids the biases of VQA/GQA setting and achieves better relative performance compared to random chance.

\section{Probing LXMERT on VSR}
We start by establishing LXMERT's image-text matching baselines on the VSR dataset by conducting the fine-tuning experiments in two different settings: (1) Fine-tuning the binary classification head only by freezing the pre-trained LXMERT weights and (2) Fine-tuning both LXMERT and its classification head end-to-end, similar to what was done in \cite{vsr}. 
\begin{table}[h]
\centering
\begin{tabular}{lcc}
\hline
\textbf{Model} & \textbf{ZS} & \textbf{Rand} \\\hline
Random Chance & 50 & 50 \\\hline
LXMERT Fine-tuning (\texttt{CLS-Head}) & 54.0 & 60.0 \\
LXMERT Fine-tuning (\texttt{End-to-End}) & \textbf{65.6} & \textbf{74.1} \\
\hline
\end{tabular}
\caption{VSR's test set performance after fine-tuning on VSR train set. \textbf{ZS} and \textbf{Rand} stand for Zero-shot and Random splits in the VSR benchmark, respectively.}
\label{tab:zeroshotVQAFull}
\end{table}
%
According to Table 1, the second setting (indicated as \texttt{End-to-End}) achieves the closest performance to what is reported in \cite{vsr}. 
This rather low performance is an indicator of the difficulty in efficient adaptation of the LXMERT-style VLMs for more complex reasoning tasks.
%
%
Other models have similar difficulties. 
More recently, authors in~\cite{herzig2023incorporating} proposed alternative pre-training of BLIP-2~\cite{blip2} with additional structured supervision provided by scene graph annotations. While this pre-training helped to achieve the accuracy of $63.4\%$ on the VSR dataset, it still underperformed the LXMERT end-to-end fine-tuning accuracy. 
Additional advancements in the pre-training objectives and data reported in~\cite{bugliarello2023weakly} using larger X-VLM model with object localization head~\cite{xvlm}, still underperforms smaller LXMERT model on VSR Random split test set.

\noindent
\textbf{Quantitative Noun Grounding Analysis.}
 Due to the straightforward grammatical structure of the captions in the VSR dataset, we can extract \textit{subject}, \textit{relationship}, and \textit{object} from the captions with the minimal effort. 
Next, we quantify the performance of the subject/object detector on the VSR dataset.
The rows marked $1.$ to $3.$ in Table 2 focus on cases when LXMERT's image-matching prediction is right, given that one, both, or none of the subject \& object is/are being detected by Faster-RCNN. 
For example, row $1.$ refers to the cases in which the binary ITM is Successful (indicated as \textbf{S}). 
but only in 27.08\% of the examples both subject and object phrases were detected category labels of Faster-RCNN outputs and 30.41\% of them were WordNet~\cite{miller1995wordnet, bird2006nltk}  synonyms (not the exact word match). 
The same applies to the Rand columns, while in row $2$, in 72.91\% of the successful binary ITM predictions, only one of the subject and object phrases were among the detections. 
\begin{table}[!h]
\small
\smallskip
\centering
\begin{tabular}{lccccc}
\hline
\textbf{Case} & \textbf{S} & \textbf{One} & \textbf{Both} & \textbf{ZS} & \textbf{Rand} \\\hline
\rowcolor{verylightgray}
$Rand$ & -- & -- & -- & 50 & 50 \\\hline
\rowcolor{verylightgreen}
$Acc$ & -- & -- & -- & 65.66 & 74.11 \\\hline
\rowcolor{verylightgreen}
$1.$ & \cmark & \cmark & \cmark & 27.08 / 30.41 & 27.93 / 32.93 \\
\rowcolor{verylightgreen}
$2.$ & \cmark & \cmark & \xmark & 72.91 / 55.41 & 72.06 / 50.73\\
\rowcolor{verylightgreen}
$3.$ & \cmark & \xmark & \xmark & 16.25 / 14.16 & 18.93 / 16.33\\\hline
\end{tabular}
\caption{LXMERT's results on original VSR test set \cite{vsr} and Faster-RCNN error analysis. ZS refers to the zero-shot setting in VSR, in which there is no concept overlap between the train/dev/test splits, while in Rand, all the data is distributed randomly. ($Acc$ stands for accuracy as \% of correctly predicted ITM binary labels.}
\label{tab:lxmertVSR}
\end{table}
A similar analysis for the incorrect image-text matching prediction can be found in Table 5 in the Appendix. 
While these numbers are not directly indicative of the ultimate ability of the model to ground subject/object nouns with ROI feature tokens, the weak correlation between them is inferred, as it was also studied in previous works like \cite{vision-for-language} and suggests challenges with subject and object grounding in VSR. 
%
Additional qualitative analysis using the explainability method of~\cite{explainability} 
can be found in Figure 5 in the Appendix. This analysis also confirms that a significant factor in understanding spatial relationships in images is the lack of noun grounding. Furthermore, the fact that image and text can be matched successfully in the absence of noun phrase grounding skews the conclusions made by ITM probing tasks.

\begin{figure*}[!t]
    \centering
     \includegraphics[width=\linewidth]{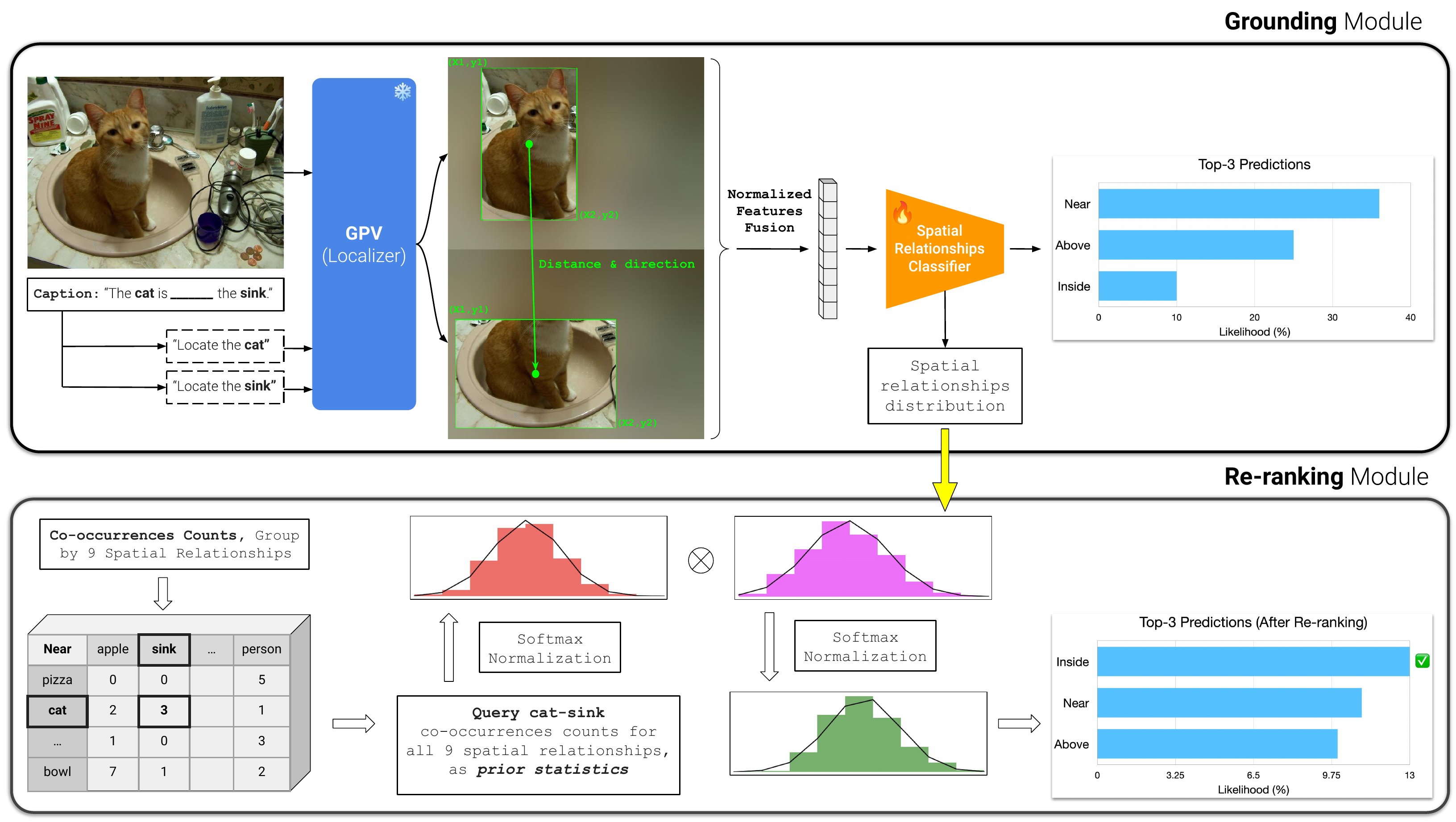}

    \caption{\textbf{Our Approach} consists of two main modules: (1) \textbf{Grounding Module} predicts the locations of objects along with their confidences, and MLP takes the 
    bounding box coordinates and predicts the distribution of spatial relationships. These are then combined to compute the initial ranking of spatial clauses. (2) \textbf{Re-ranking Module} adjusts the ranking given the co-occurrence priors. This example shows the effectiveness of the Re-ranking Module in adjusting the spatial clause distribution (which brings $inside$ to the 1st rank), while the initial top-3 predictions were semantically correct, anyway.}
    \label{fig:fig3}
\end{figure*}

\section{Approach}
\label{sec:approach}
We propose a \textit{\textbf{compositional}} approach for spatial reasoning by \textit{\textbf{decoupling}} the process of (1) grounding \texttt{Subject}, and \texttt{Object} of the <\texttt{Subject}, \texttt{Relationship}, \texttt{Object}> triplet extracted from the ground-truth caption and (2) predicting the \texttt{Relationship} by training a multi-class classification model using location features obtained by the detector. This approach is motivated by earlier structured Conditional Random Fields (CRF) methods for modeling image content using scene graphs, like in \cite{crf} that applied it to image retrieval tasks. 
We used encoder-decoder vision-language models, MDETR \cite{mdetr} and GPV \cite{gpv}, and their ability to query the locations of objects. Our approach is outlined in Figure 2.


\noindent\textbf{Grounding Module} takes as input an image and spatial clause in the form of <\texttt{Subject}, \texttt{Relation}, \texttt{Object}> and queries $GPV_{Localizer}$ with the prompt of $Q_i = \mbox{"\textit{Locate the Subject}"}$, getting back normalized bounding box coordinates as $l_i =[x_i, y_i, h_i, w_i]$ and confidence score $p(i) = Pr(o_i | I, Q_i)$ of the most confident prediction given the query $Q_i$.
Similarly for $p(j) = Pr(o_j | I, Q_j)$ with $Q_j = \mbox{"\textit{Locate the Object}"}$. The concatenation of the bounding box coordinates is fed to the $MLP$ to generate the initial probability distribution over all $r_k$ spatial relationships\footnote{$k=9$ in our case.}.
$\left\{Pr_{ij}^1, ..., Pr_{ij}^{k} \right\} = MLP ([l_i,l_j])$, where $Pr_{ij}^k = Pr(r_k|l_i, l_j)$.
The score of the spatial clause is then calculated using a simple scoring function where score $S_k(I,T)$ is computed as follows:
\\
\[S_k(I,T) = p(i) Pr(r_k|l_i, l_j) p(j). \]
\\
\noindent\textbf{Re-ranking Module} then uses the prior probability distribution $Pr[R_k(i, j)]$ of two objects $i$ and $j$ appearing in certain spatial relation $k$, computed by 
counting all prior co-occurrences of $i$ and $j$ appearing in relation $k$.
The final relation probability is then ranked by 
$r_k(i,j) = Pr(r_k|l_i, l_j)Pr[R_k(i, j)] $ yielding the final ranking function:
\\
\[S_k(I,T) = p(i) r_k(i,j) p(j). \]

\section{Experiments}

\label{sec:dataPrep}
In terms of data pre-processing, we first discarded the orientation-based clauses like \textit{"facing away"} and \textit{"parallel to"},  which require the pose or 3D information. VSR also reported this class of spatial clauses as the worst-performing one, even below the random chance, and the depth-based clauses like \textit{"in front of"}, \textit{"behind"},  that require depth signal. Then, we grouped the semantically similar spatial clauses in VSR together and ended up with nine merged classes as: (1) \textit{below}, (2) \textit{above}, (3) \textit{far from}, (4) \textit{right of}, (5) \textit{left of}, (6) \textit{inside}, (7) \textit{outside}, (8) \textit{near}, and (9) \textit{contains}, to have more controlled experiments over primitive spatial clauses (details of spatial clauses merging/grouping can be found in Table 6). In total, we ended up with 3895 data samples, which we divided randomly into 3116 (80\%) and 779 (20\%) splits using \texttt{StratifiedSampling} to be used as the gold train and test sets for our experiments. We also experimented with feeding additional geometric features (normalized relative distance \& direction from the center of subject towards the center of object) and data augmentation (by swapping subjects and objects for symmetric clauses, as well as swapping + negating asymmetric spatial clauses) to the MLP for spatial relationship classification. 
%
%
\begin{table}[!h]
\centering
\begin{tabular}{lcc}
\hline
\textbf{Model} & \textbf{Top-1} & \textbf{Top-3}\\\hline
Random Chance$_{Multiclass}$ & 11.11 & 33.33 \\
\hdashline
Bbox w/o Re-ranking & 41.97 & 83.05 \\
\hspace{1mm} + $geo$ & 40.17 & 82.54 \\ 
\hspace{1mm} + $geo$ + $aug$ & 42.36 & 80.61 \\ 
\hdashline
Bbox w/ Re-ranking & 53.78 & \textbf{88.96}\\
\hspace{1mm} + $geo$ & \textbf{54.04} & 88.31 \\ 
\hspace{1mm} + $geo$ + $aug$ & \textbf{54.04} & 87.54 \\\hline
\textbf{Ours Best $\Delta$} & \textbf{42.93} & \textbf{55.63} \\\hline
\end{tabular}
\caption{Ablation study of our approach results on our test set, comparing to the random chance performance similar to the \cite{winoground} evaluation style. "$geo$" means adding three geometry-based features to MLP, and "$aug$" means data augmentation for our training split of bounding box locations.}
\label{tab:ourResults}
\end{table}

\noindent
\textbf{Results Analysis.} Table 3 shows the performance of our ranking approach, trained and tested on a subset of VSR.
Our top-performing settings of 54.04\% and 88.96\% perform 42.93\% and 55.63\% above the random chances reflecting the \% of times the correct spatial relationship is the \textit{top-1} or among the \textit{top-3} ranked spatial clauses. \\
\noindent
\textbf{VQA/GQA Baselines Comparison}
In order to have a fair comparison on the VSR benchmark of the state-of-the-art VLM's with different architectures and pre-training strategies we chose: (1) LXMERT from \textit{Cross-Modal}, (2) MDETR from \textit{Modulated}, and (3) GPV from \textit{Encoder-Decoder} Transformers categories. Since VSR benchmark is curated as image-text pairs with binary labels for ITM, we modified the image-text matching into a "yes/no" VQA/GQA task by turning the captions into questions (using the same triplet information) and considering "yes" as \textbf{1} and "no" as \textbf{0} binary labels, so we can test models that do not have the ITM classification head in place. 
The results of golden test set (both for Zero-shot and Random split of VSR) and  are shown in Table 4.
\begin{table}[!h]
\centering
\begin{tabular}{lc|ccc}
\hline
\textbf{Model} & \textbf{Binary} & & \textbf{ZS} & \textbf{Random} \\\hline
Random Chance$_{Binary}$ & 50 &  & 50 & 50  \\
GPV VQA \cite{gpv} & 58.27 & \mbox{} &52.25 & 54.69 \\
MDETR GQA \cite{mdetr} & \textbf{95.63} & \mbox{} & 53.35 & 54.54 \\
LXMERT VQA \cite{tan2019lxmert} & 62.38 & \mbox{} & - & - \\
LXMERT GQA \cite{tan2019lxmert} & 95.37 & \mbox{} & \textbf{57.71} & \textbf{55.23} \\\hline
\textbf{SOTA Best $\Delta$} & \textbf{45.63} & \mbox{} & \textbf{7.71} & \textbf{5.23}\\\hline
\end{tabular}
\caption{Pre-trained SOTA results on our test set (indicated as \textbf{Binary} on the left) vs. on VSR test set (indicated as \textbf{ZS} and \textbf{Random} on the right)}
\label{tab:SOTAResults}
\end{table}
Although these models (specifically their GQA classification heads) are already fine-tuned on multiple V+L tasks, 
they still perform slightly above the random chance in the zero-shot inference. 
MDETR \textit{pre-trained} on GQA \cite{hudson2019gqa} achieves the highest performance on our test set, performing 45.63\% above the binary random chance. Note that the biggest improvement was marked by models pre-trained on GQA (MDETR GQA and LXMERT GQA). We hypothesize that GQA contains more spatial clauses that resemble our modified dataset. Comparing our best performance after re-ranking (showed as \textbf{Ours Best $\Delta$ = 55.63} in Table 3) to the best performing model of the baselines (showed as \textbf{SOTA Best $\Delta$ = 45.63} on Table 4), shows that our approach outperforms the best SOTA (MDETR GQA) by 10\%, in terms of the relative accuracy over the random chance, on the same test set.

\section{Related Works}
Recent years have witnessed a fast-paced emergence of different multi-modal vision and language architectures and pretraining strategies. We first review the representative vision-language models and briefly discuss their strengths, weaknesses, and limitations pertaining to the study of the visual spatial reasoning task. \\

\noindent\textbf{Cross-Modal Encoder Transformers} follow the architecture of ViLBERT \cite{lu2019vilbert}, LXMERT \cite{tan2019lxmert}, UNITER \cite{uniter}, OSCAR \cite{oscar}, and VinVL \cite{vinvl}.
The visual input in these models is typically tokenized using object proposals and their associated region of interest features (and bounding box coordinates) obtained by pretrained object detectors such as Faster-RCNN. While these models achieve impressive performance using smaller amounts of training data ($\sim$9.7 million image-text pairs) for pretraining, they are not end-to-end trainable, and their performance on downstream tasks is affected by the quality of the detected regions.

Several probing studies using specially curated datasets demonstrated that these models lack the understanding of attribution, relations, order, verbs, and counting~\cite{stanford-bag-of-words, svo, winoground, kamath2023s}. 
Since LXMERT was the best-performing model on the VSR \cite{vsr} benchmark, we include this model in our experiments as the representative baseline from this class of architectures. \\

\noindent\textbf{Dual-Encoder Transformers} were first introduced by CLIP~\cite{clip} use contrastive learning and a large dataset of image-text pairs ($\sim$400 million) and to align holistic image and text representations, as well as ALIGN \cite{align} (using $\sim$1.8 billion image-text pairs). While CLIP demonstrated high performance on image, scene, and action classification benchmarks, multiple probing studies of fine-grained understanding have demonstrated poor performance on tasks that require more compositional reasoning. For example, ~\cite{reclip} showed on the CLEVR dataset that if a spatial clause is added to the caption, the performance of image-text matching is at the level of random chance. Follow-up works combine using holistic representations of Dual-Encoders with tokenized representations of Cross-Modal Encoders such as in ALBEF~\cite{albef}, or adding single modality pretraining to this new architecture like FLAVA~\cite{flava}. The lack of compositionality and fine-grained understanding still remains, according to the findings of \cite{winoground}. \\

\noindent\textbf{Modulated Transformers} are trained end-to-end and can predict bounding box coordinates associated with noun phrases in the text description but require correspondences between bounding boxes and noun phrases for the training stage. 
MDETR \cite{mdetr} is representative of this category, which is built on top of DETR \cite{detr} and trained for fine-grained alignment between regions of interest in the image and associated noun phrases using the Hungarian matching algorithm. Additional pre-training tasks include VQA and GQA with their own classification heads. GLIP \cite{glip} follows a similar approach but uses a different architecture using a series of cross-attention layers from a language encoder and transformer-based object detection module. 
Due to the specific capabilities of MDETR (i.e. grounding and pre-trained GQA head), we used this model as another baseline for our experiments.\\

\noindent\textbf{Encoder-Decoder Transformers} were introduced by the General Purpose Vision (GPV) \cite{gpv} model. This model has object localization functionality, can do image classification, captioning, and VQA tasks. The task is specified as input to the model in natural language, enabling simultaneous multi-task training. The noun phrase localization task has bounding box ground truth available during training, while other tasks such as classification, question answering, and captioning have ground truth text associated with images. Other task-agnostic vision-language models that are end-to-end trainable on multiple tasks
and have generative decoder branches for auto-regressing text include~\cite{blip, simvlm, coca}. 
The localization and VQA capabilities of GPV make this model suitable for our experiments. We built our ranking model on top of the GPV Localization module, which is demonstrated in Figure 2.

\noindent\textbf{Spatial Relationship Understanding.} Previous works on spatial relationship understanding use the synthetic CLEVR \cite{johnson2017clevr} dataset, focus on simpler spatial relationships and neglect the challenges in real images posed by object detection and representation learning. More general approaches study spatial relationships as part of VQA \cite{vqa} or GQA \cite{hudson2019gqa} tasks. 
According to the GQA benchmark, only 8\% and 22.4\% of VQA and GQA are allocated to \textit{Spatial Reasoning} questions, respectively. Also, the existing GQA questions that probe spatial understanding typically have binary YES/NO answers and 
hence inherit the well-known biases of the VQA/GQA task. VSR \cite{vsr} is the most recent dataset curated for studying spatial clause understanding in a more visually realistic setting using MSCOCO \cite{mscoco} images. 
The authors report performance on image-text matching tasks by fine-tuning the existing models (LXMERT, VisualBERT, and ViLT) both in zero-shot setting and after fine-tuning.
\section{Conclusions and Future Works}
In this work, we demonstrated the low zero-shot performance of several state-of-the-art VLMs on spatial reasoning task and demonstrated that this is partly caused by the lack of grounding of noun phrases in models pre-trained with image-text matching using holistic representations of each modality, like in LXMERT \cite{tan2019lxmert}. Also, we proposed a compositional approach for spatial reasoning using outputs of the GPV encoder-decoder model with explicit quantification of subject and object grounding and spatial relationship classification/ranking. This approach outperformed the SOTA models in terms of the relative increase over the random chance for each setting (a similar evaluation strategy used in \cite{winoground} and \cite{kamath2023whatsupvlms}). Another major advantage of our approach is the modularity that makes it possible to replace the localization module with any upcoming SOTA Vision-and-Language model (with the grounding capability) in the future and just re-train the lightweight spatial relationship classification head. However, in the presented approach, the reasoning is done purely in 2D with decoupled locations and region ROI features. Further disambiguation of more complex relations requires knowledge of 3D, which can be either revealed by depth and object pose estimates or novel larger datasets, additional means of supervision, or variations in model architectures. Finally, additional drawback of the VSR dataset (which should be considered during the generation of future datasets for studying spatial reasoning) is the lack of ground-truth bounding box annotations of subject and object. In this work, we considered the GPV localization \texttt{relevance scores} as silver labels (as the majority of the scores fall between 0.8 and 1.0 for the entire dataset, and their distribution histogram can be found in Figure 7 in the Appendix).

\vskip 0.2in
\bibliography{sample}

\appendix
\section{}
\label{app:theorem}







\subsection{Implementations Details}
For the LXMERT reproduction results, we used the \texttt{huggingface} \cite{wolf2019huggingface} checkpoint for the first setting and VSR codebase for the second setting, respectively. For the first setting, pass the 768-dim \texttt{Cross-Modality CLS Token} into a 3-layer MLP with the size of (768, 512, 64, 1), followed by \texttt{sigmoid} activation for binary classification. Then, the model is trained for 200 epochs and saved the checkpoint based on the early stopping of the validation loss. For the localization part of our model, we used \texttt{GPV-1-Loc} checkpoint, pre-trained on \texttt{COCO} split. For spatial relationship reasoning, our MLP model consists of (1) an input size of 8, (2) two hidden layers of 16 and 32, and (3) an output layer with nine neurons, respectively. Both hidden layers are followed by \texttt{BatchNorm1d} layers, then \texttt{ReLU} activation. Training has been done for 100 \texttt{epochs} until the convergence, with the \texttt{batch-size} of 12 and \texttt{learning-rate} of 1e-5, using the \texttt{CrossEntropyLoss} criterion and \texttt{Adam} optimizer. All the above scripts are mainly developed using \texttt{Pytorch} and \texttt{transformers} libraries.

\subsection{LXMERT Zero-shot vs. Fine-Tuning Settings}
Provided in Figure 3.  

\begin{figure}[h]
    \centering
    \includegraphics[scale=0.38]{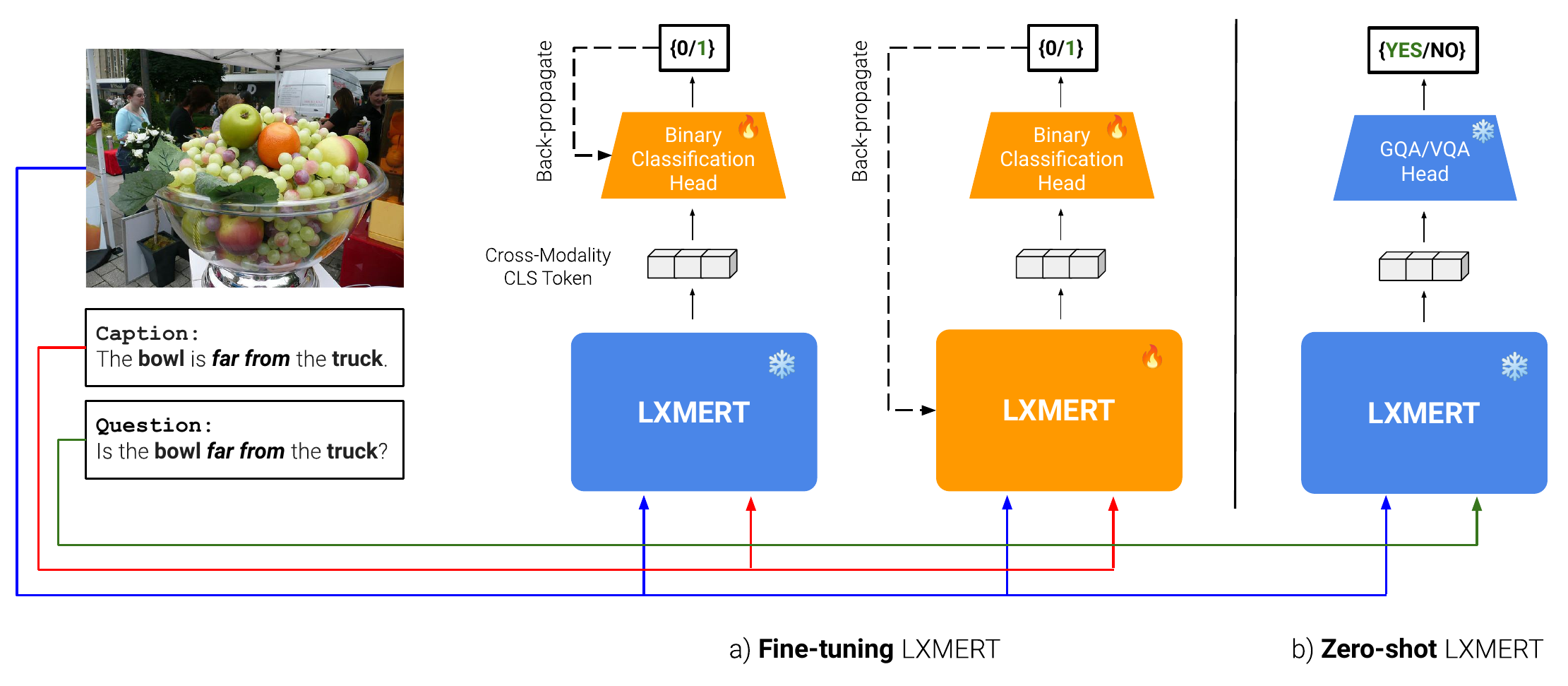}
    \caption{LXMERT Fine-tuning and Zero-shot Performance Reproduction Experiments Design}
    \label{fig:sub1}
\end{figure}

\subsection{Learning Curves of Fine-tuning LXMERT on VSR (CLS-Head Only)}
Provided in Figure 4.

\begin{figure*}[!h]
\centering
        \begin{subfigure}[b]{1.0\textwidth}
                \centering
                \includegraphics[width=\linewidth]{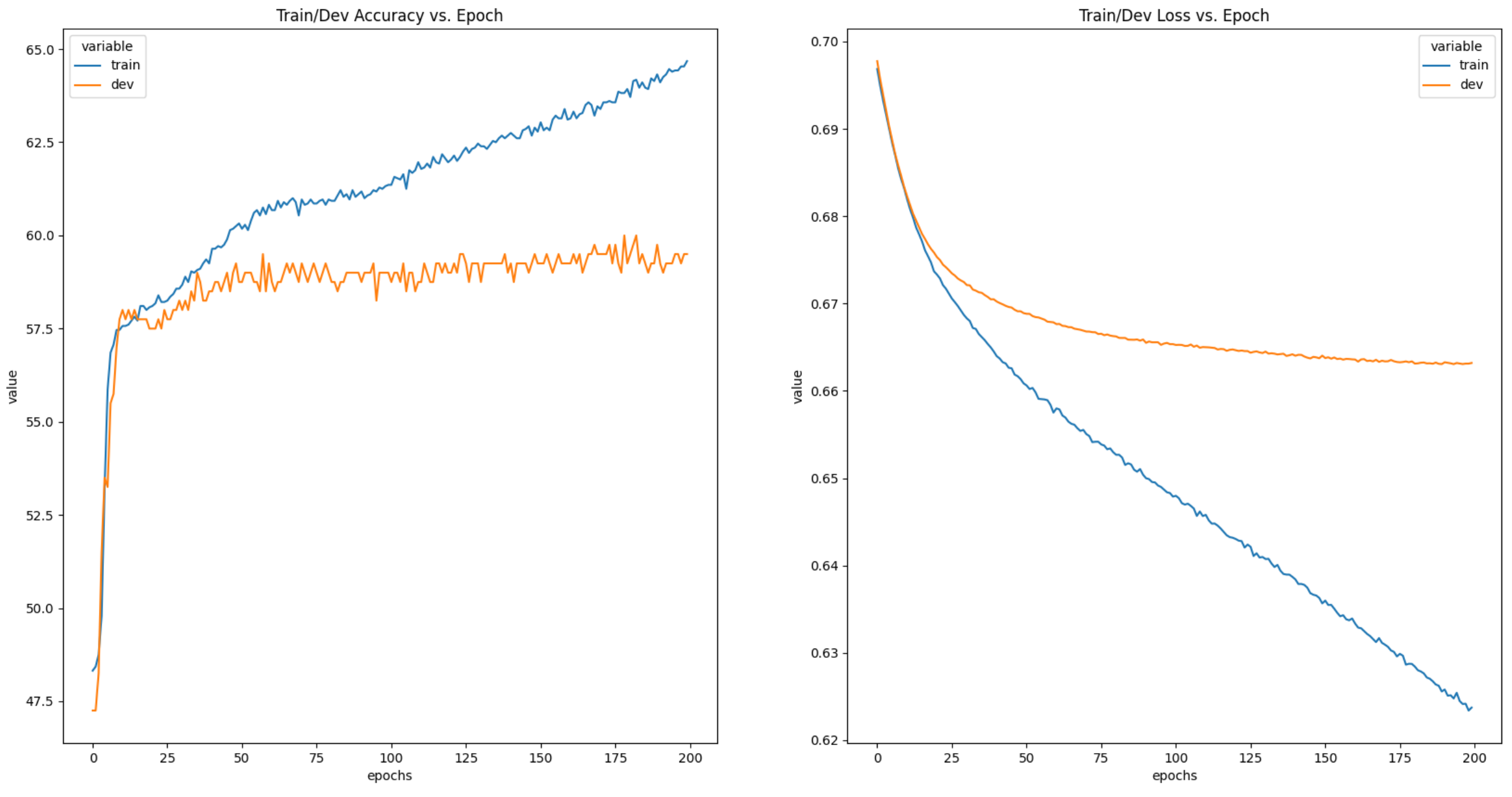}
                \caption{Random Split}
                \label{fig:subfig1}
        \end{subfigure}\hfill
        \vspace{1cm}
        \begin{subfigure}[b]{1.0\textwidth}
                \centering
                \includegraphics[width=\linewidth]{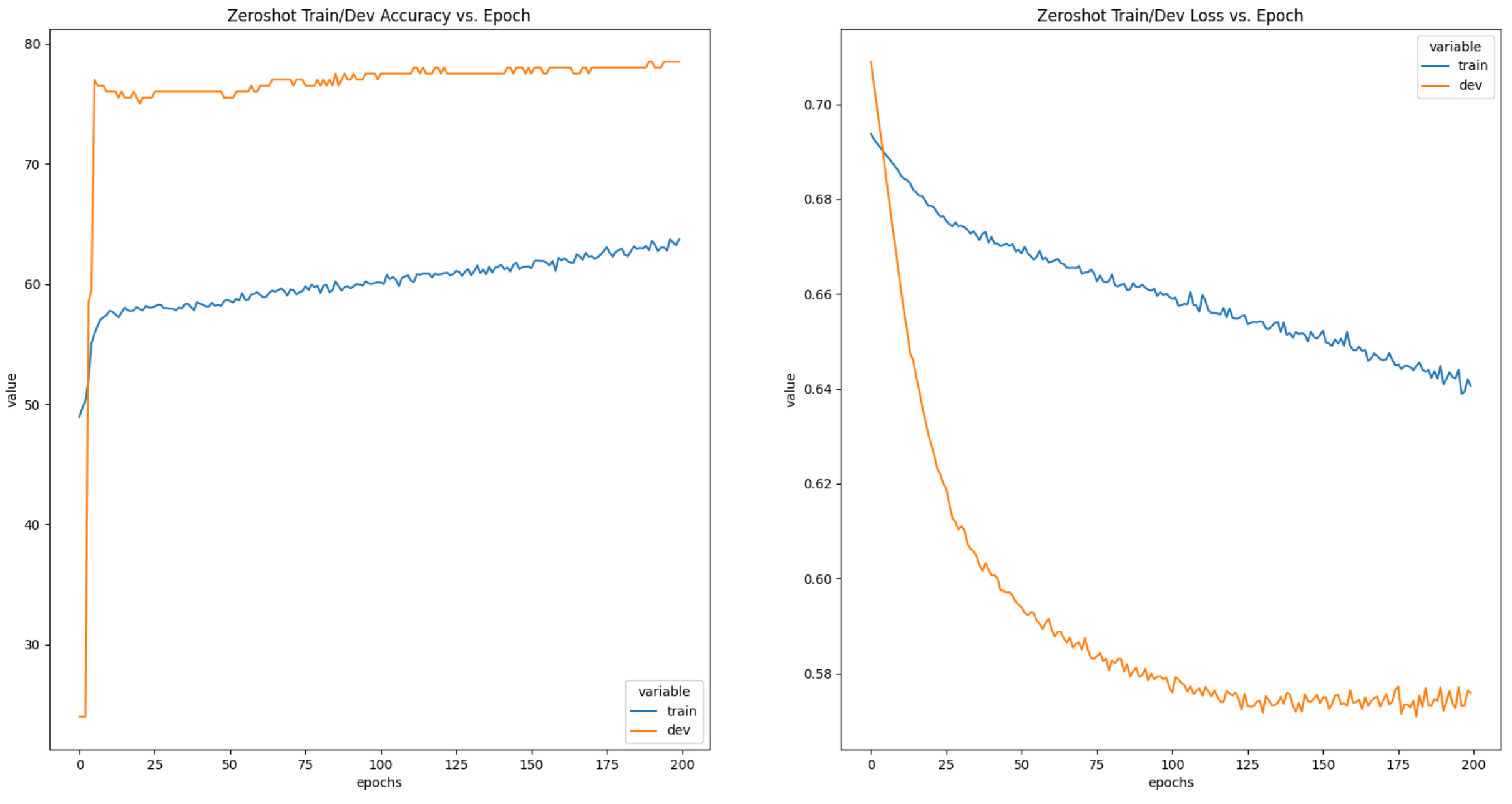}
                \caption{Zero-shot Split}
                \label{fig:subfig2}
        \end{subfigure}\hfill
        \caption{Training curves for LXMERT Binary Classification Head Fine-tuning on VSR Train Set}
        \label{fig:lxmertBinaryHeadCurves}
\end{figure*}

\subsection{Full LXMERT Quantitative Results}
Provided in Table 5.

\begin{table}[h]
\small
\smallskip
\centering
\begin{tabular}{lccccc}
\hline
\textbf{Case} & \textbf{S} & \textbf{One} & \textbf{Both} & \textbf{ZS} & \textbf{Rand} \\\hline
\rowcolor{verylightgray}
$Rand$ & -- & -- & -- & 50 & 50 \\\hline
\rowcolor{verylightgreen}
$Acc$ & -- & -- & -- & 65.66 & 74.11 \\\hline
\rowcolor{verylightgreen}
$1.$ & \cmark & \cmark & \cmark & 27.08 / 30.41 & 27.93 / 32.93 \\
\rowcolor{verylightgreen}
$2.$ & \cmark & \cmark & \xmark & 72.91 / 55.41 & 72.06 / 50.73\\
\rowcolor{verylightgreen}
$3.$ & \cmark & \xmark & \xmark & 16.25 / 14.16 & 18.93 / 16.33\\\hline
\rowcolor{verylightred}
1-$Acc$ & -- & -- & -- & 34.34 & 25.89 \\\hline
\rowcolor{verylightred}
$4.$ & \xmark & \cmark & \cmark & 29.48 / 34.26 &  25.19 / 28.43 \\
\rowcolor{verylightred}
$5.$ & \xmark & \cmark & \xmark & 70.51 / 54.98 &  74.80 / 52.86 \\
\rowcolor{verylightred}
$6.$ & \xmark & \xmark & \xmark & 11.95 / 10.75 &  20.61 / 18.70 \\\hline
\end{tabular}
\caption{LXMERT's results on original VSR test set \cite{vsr} and Faster-RCNN error analysis. ZS refers to the zero-shot setting in VSR, in which there is no concept overlap between the train/dev/test splits, while in Rand, all the data is distributed randomly. ($Acc$ stands for accuracy as \% of correctly predicted ITM binary labels.}
\label{tab:lxmertVSR}
\end{table}

\subsection{LXMERT Explainability Results Visualization}
Provided in Figure 5. Four different cases for which LXMERT predicts image-text labels successfully in all of them. The first column shows the original image-text pairs from the VSR dataset. The second and third columns demonstrate the regions with the highest relevancy score computed from attention weights using~\cite{explainability}, for \textit{subject} and \textit{object}, respectively.
\label{sec:explainability}

\begin{figure*}
\centering

        
        \begin{subfigure}[t]{0.3\textwidth}
                \centering
                \includegraphics[width=\linewidth]{figures/cup_hotdog.jpg}
                \captionsetup{justification=centering, margin=0.5cm}
                \caption{The cup is to the left of the hot dog.}
                \label{fig:subfig1}
        \end{subfigure}\hfill
        \begin{subfigure}[t]{0.3\textwidth}
                \centering
                \includegraphics[width=\linewidth]{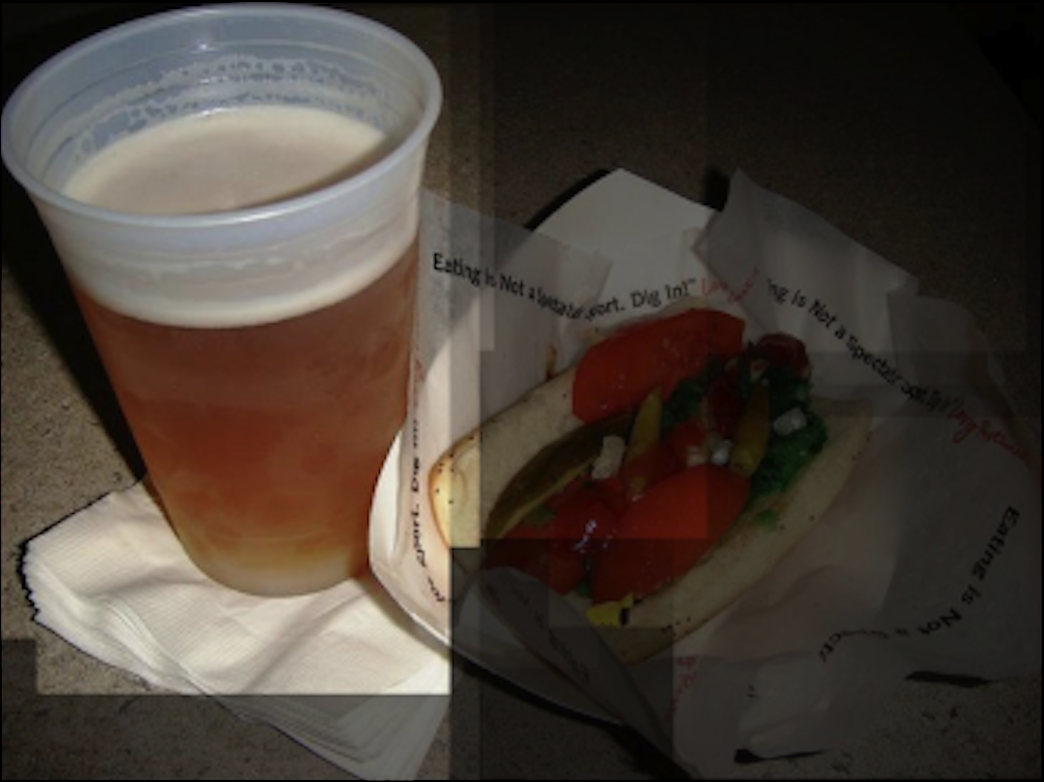}
                \caption{Locate the \textbf{\underline{cup}}}
                \label{fig:subfig2}
        \end{subfigure}\hfill
        \begin{subfigure}[t]{0.3\textwidth}
                \centering
                \includegraphics[width=\linewidth]{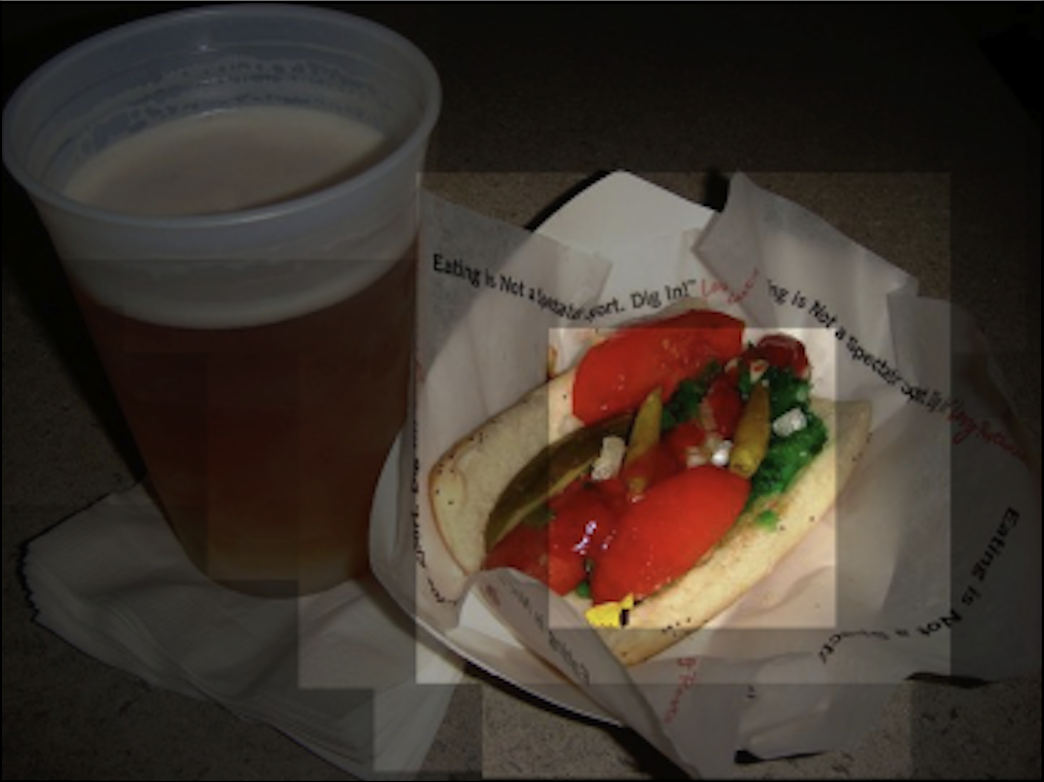}
                \caption{Locate the \textbf{\underline{hot dog}}}
                \label{fig:subfig2}
        \end{subfigure}\hfill

        
        \begin{subfigure}[t]{0.3\textwidth}
                \centering
                \includegraphics[width=\linewidth]{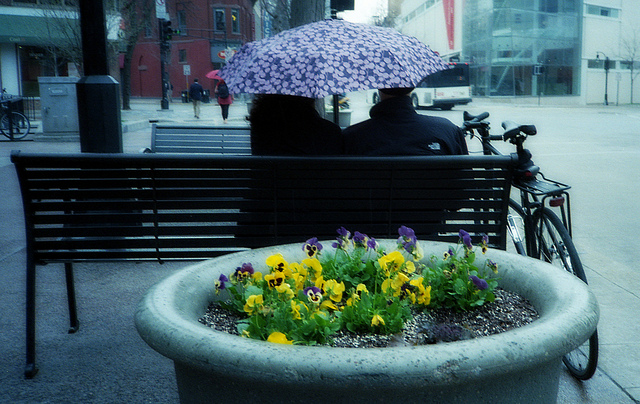}
                \captionsetup{justification=centering, margin=0.5cm}
                \caption{The potted plant is far away from the bus.}
                \label{fig:subfig3}
        \end{subfigure}\hfill
        \begin{subfigure}[t]{0.3\textwidth}
                \centering
                \includegraphics[width=\linewidth]{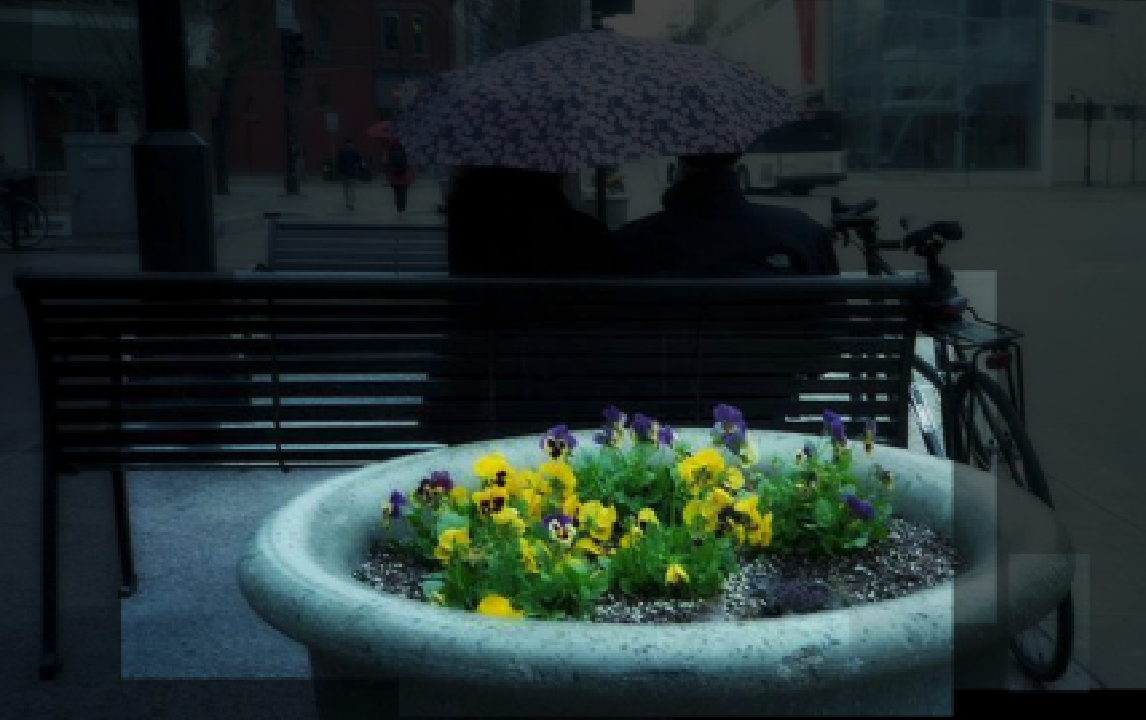}
                \caption{Locate the \textbf{\underline{potted plant}}}
                \label{fig:subfig4}
        \end{subfigure}\hfill
        \begin{subfigure}[t]{0.3\textwidth}
                \centering
                \includegraphics[width=\linewidth]{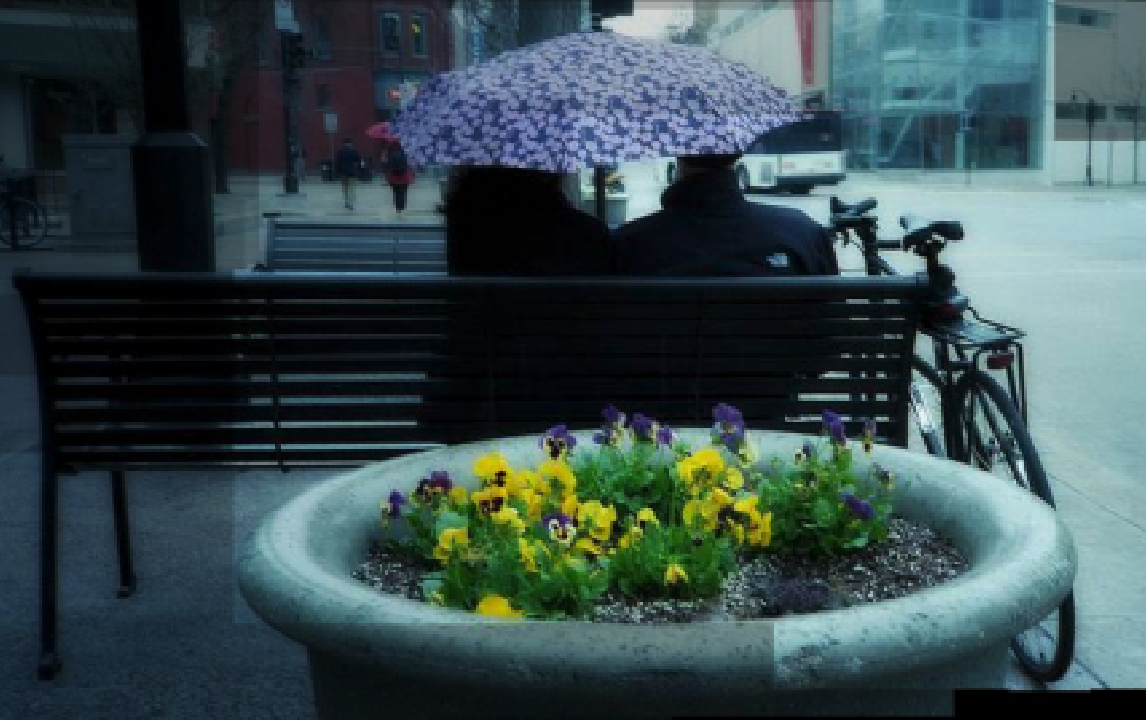}
                \caption{Locate the \textbf{\underline{bus}}}
                \label{fig:subfig4}
        \end{subfigure}\hfill

        
        \begin{subfigure}[t]{0.3\textwidth}
                \centering
                \includegraphics[width=\linewidth]{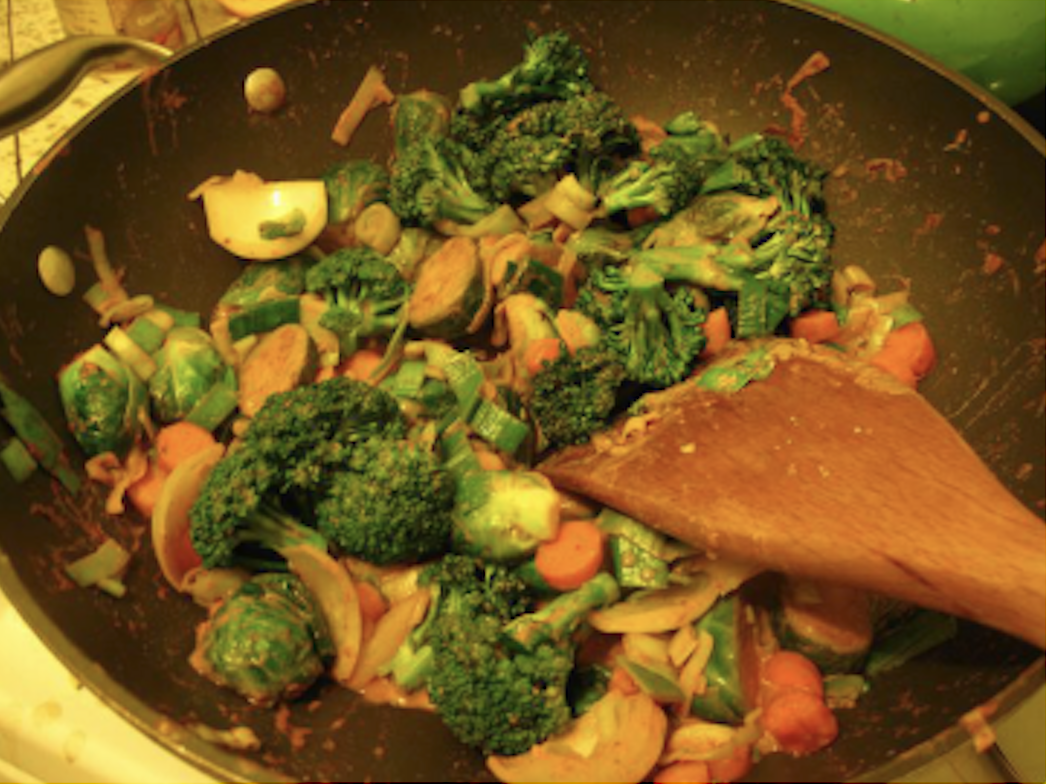}
                \captionsetup{justification=centering, margin=0.5cm}
                \caption{The bowl is beneath the spoon.}
                \label{fig:subfig1}
        \end{subfigure}\hfill
        \begin{subfigure}[t]{0.3\textwidth}
                \centering
                \includegraphics[width=\linewidth]{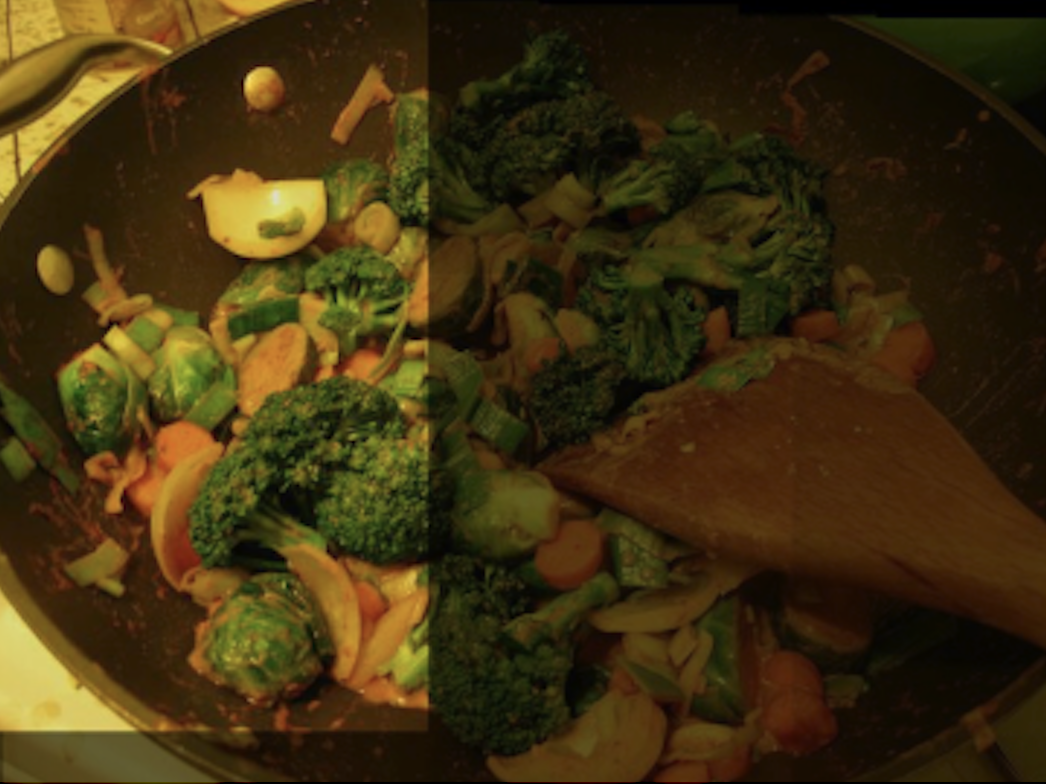}
                \caption{Locate the \textbf{\underline{bowl}}}
                \label{fig:subfig2}
        \end{subfigure}\hfill
        \begin{subfigure}[t]{0.3\textwidth}
                \centering
                \includegraphics[width=\linewidth]{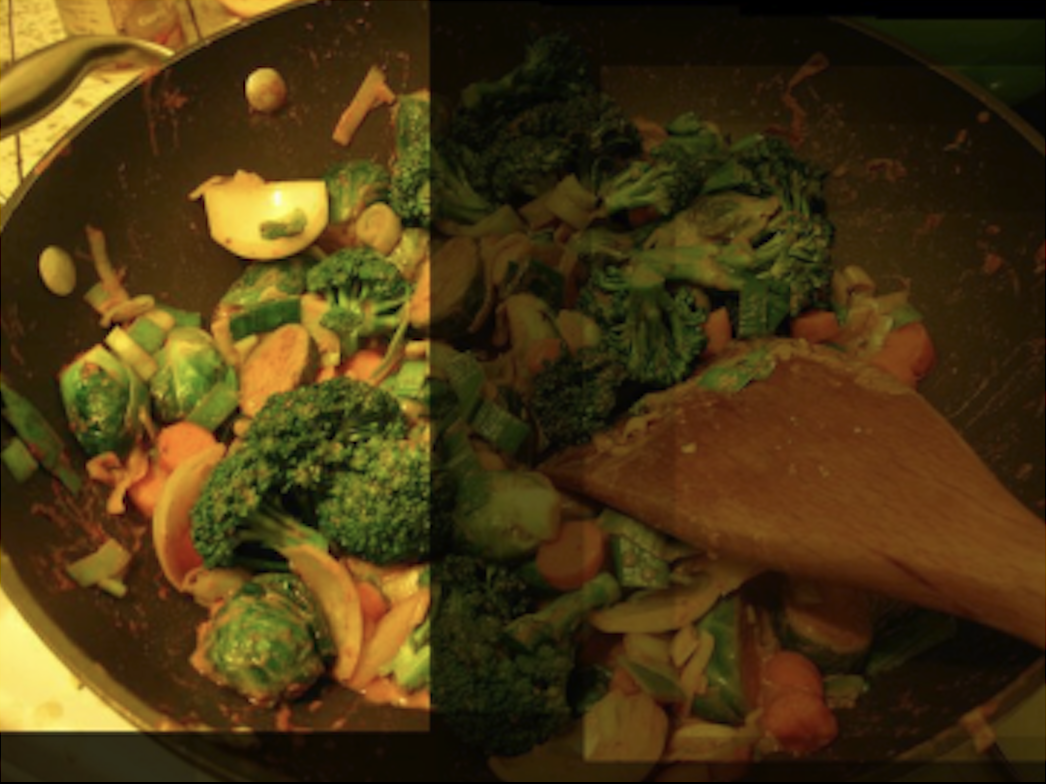}
                \caption{Locate the \textbf{\underline{spoon}}}
                \label{fig:subfig2}
        \end{subfigure}


        \begin{subfigure}[t]{0.30\textwidth}
            \centering
            \includegraphics[scale=0.3]{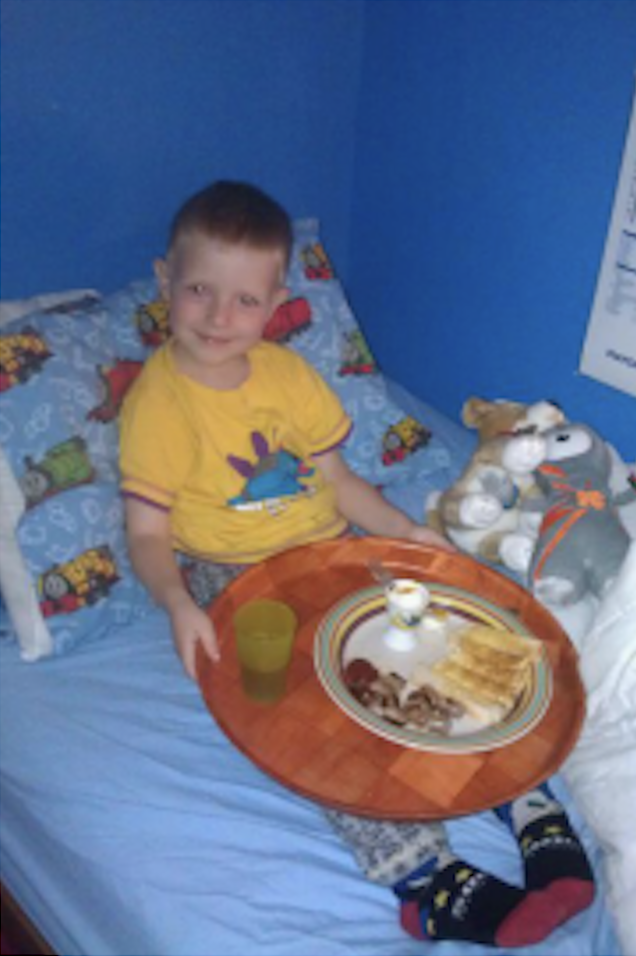}
            \captionsetup{justification=centering, margin=0.5cm}
            \caption{The cup is touching the bed.}
            \label{fig:subfig3}
        \end{subfigure}\hfill
        \begin{subfigure}[t]{0.30\textwidth}
            \centering
            \includegraphics[scale=0.3]{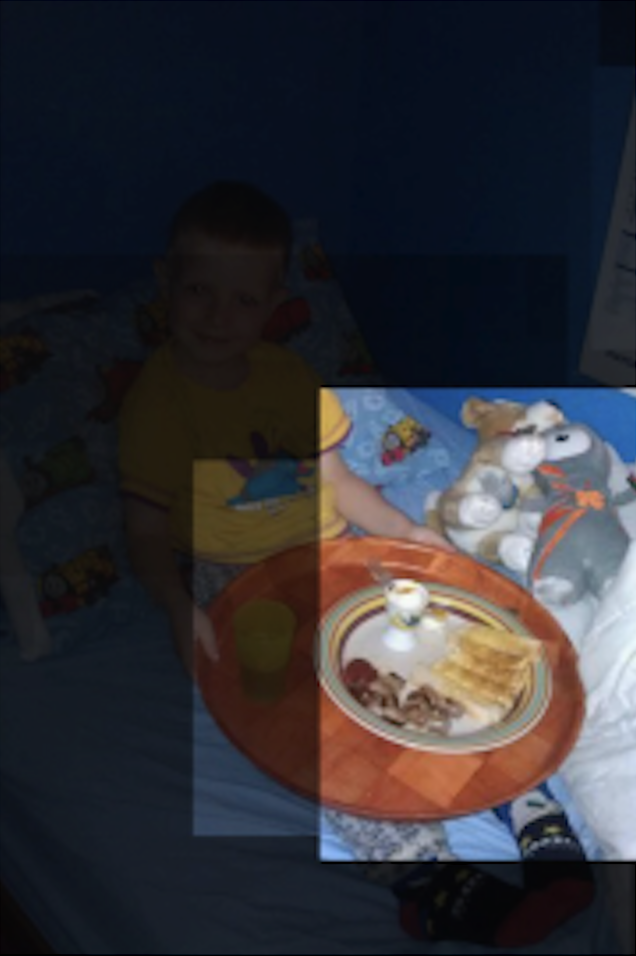}
            \caption{Locate the \textbf{\underline{cup}}}
            \label{fig:subfig4}
        \end{subfigure}\hfill
        \begin{subfigure}[t]{0.30\textwidth}
            \centering
            \includegraphics[scale=0.3]{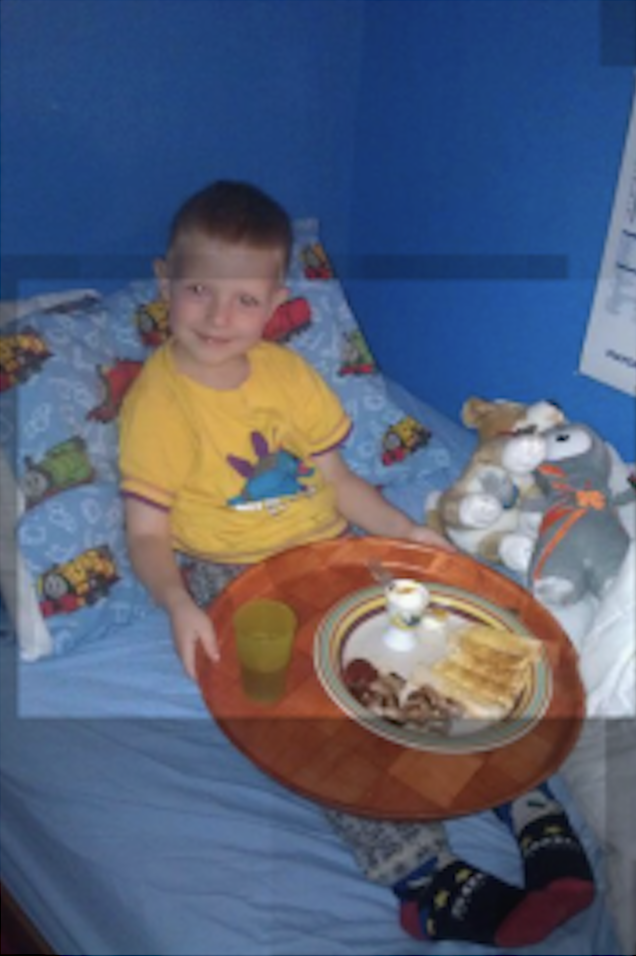}
            \caption{Locate the \textbf{\underline{bed}}}
            \label{fig:subfig4}
        \end{subfigure}\hfill

        \caption{\textbf{LXMERT Relevancy Scores}: The first row shows an example that both \textit{subject} and \textit{object} attentions imply successful grounding. The second row demonstrates relevant activations for the subject (\textit{potted plant}) but irrelevant attention weights for \textit{bus}. However, the third and fourth rows depict irrelevant attention weights for both subjects and objects, demonstrating inconsistency in LXMERT's fine-grained grounding while predicting the binary labels correctly.} 


        \label{fig:explainability}

\end{figure*}

\subsection{Spatial Relationship Groupings}
Provided in Table 6.
\begin{table}[!h]
    \centering
    \small
    \begin{tabular}{|l|p{0.7\columnwidth}|} 
    \hline
        \textbf{Merged} & \textbf{Original} \\ \hline
        \rowcolor{Gray}
        below & \textit{below}, \textit{beneath}, \textit{under}\\ \hline
        above & \textit{above}, \textit{on}, \textit{on top of}, \textit{over} \\ \hline
        \rowcolor{Gray}   
        far from & \textit{away from}, \textit{far away from}, \textit{far from} \\ \hline    
        right of & \textit{at the right side of}, \textit{right of} \\ \hline
        \rowcolor{Gray}
        left of & \textit{at the left side of}, \textit{left of} \\ \hline
        inside & \textit{in}, \textit{in the middle of}, \textit{inside}, \textit{part of}, \textit{within} \\ \hline
        \rowcolor{Gray}
        outside & \textit{outside} \\ \hline
        near & \textit{adjacent to},
              \textit{at the edge of},
              \textit{at the side of},
              \textit{attached to},
              \textit{beside},
              \textit{by}, 
              \textit{close to},
              \textit{connected to},
              \textit{near},
              \textit{next to},
              \textit{touching} \\ \hline
        \rowcolor{Gray}
        contains & \textit{contains} \\ \hline
    \end{tabular}
    \caption{Our spatial clauses merging/mapping}
\end{table}

\subsection{Confusion Matrix of Spatial Relations Classifier}
Provided in Figure 6.
\label{sec:confusion}

\begin{figure*}[!h]
\centering
        \begin{subfigure}[b]{0.45\textwidth}
                \centering
                \includegraphics[scale=0.11]{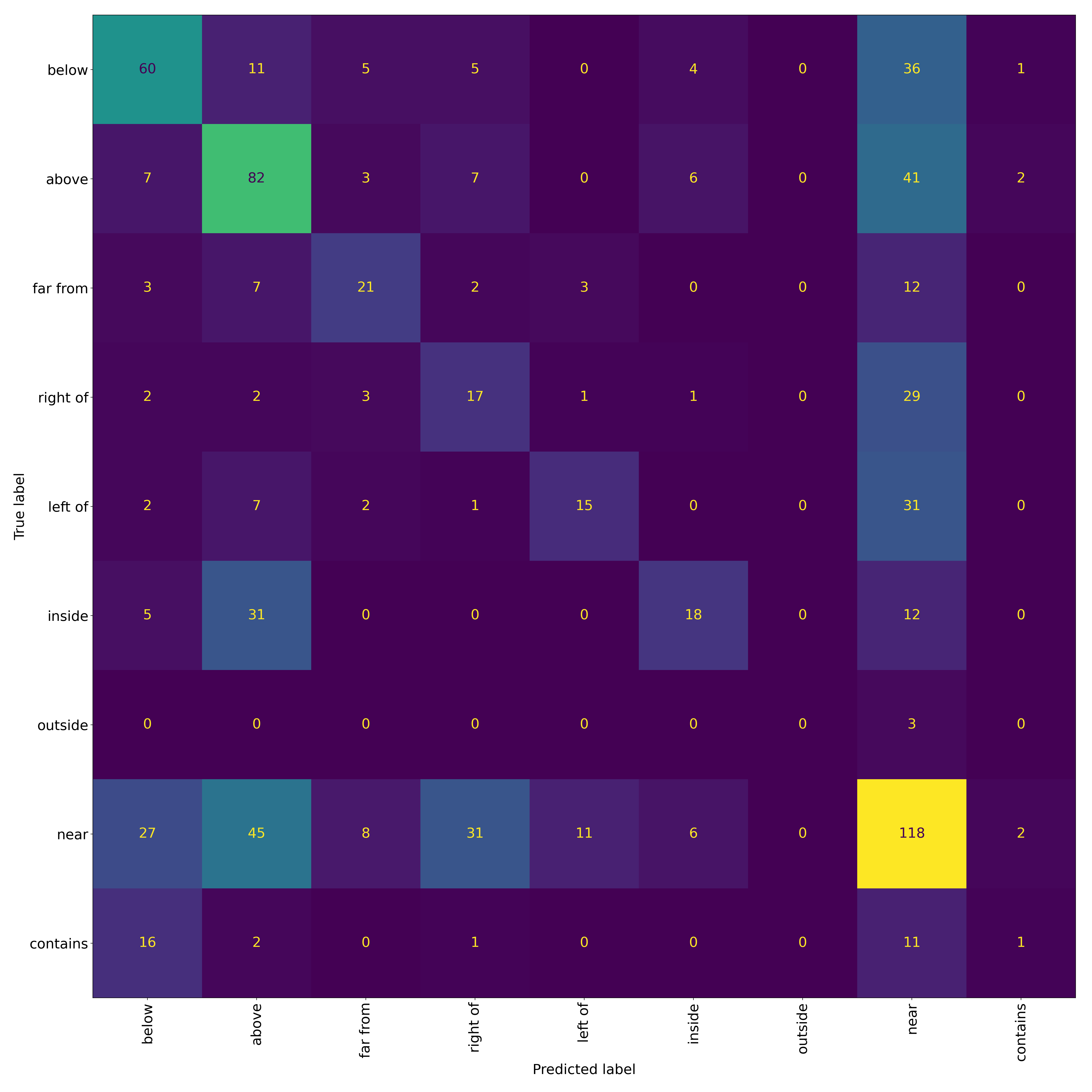}
                \caption{\textbf{Before} re-ranking}
                \label{fig:subfig1}
        \end{subfigure}\hfill
        \begin{subfigure}[b]{0.45\textwidth}
                \centering
                \includegraphics[scale=0.11]{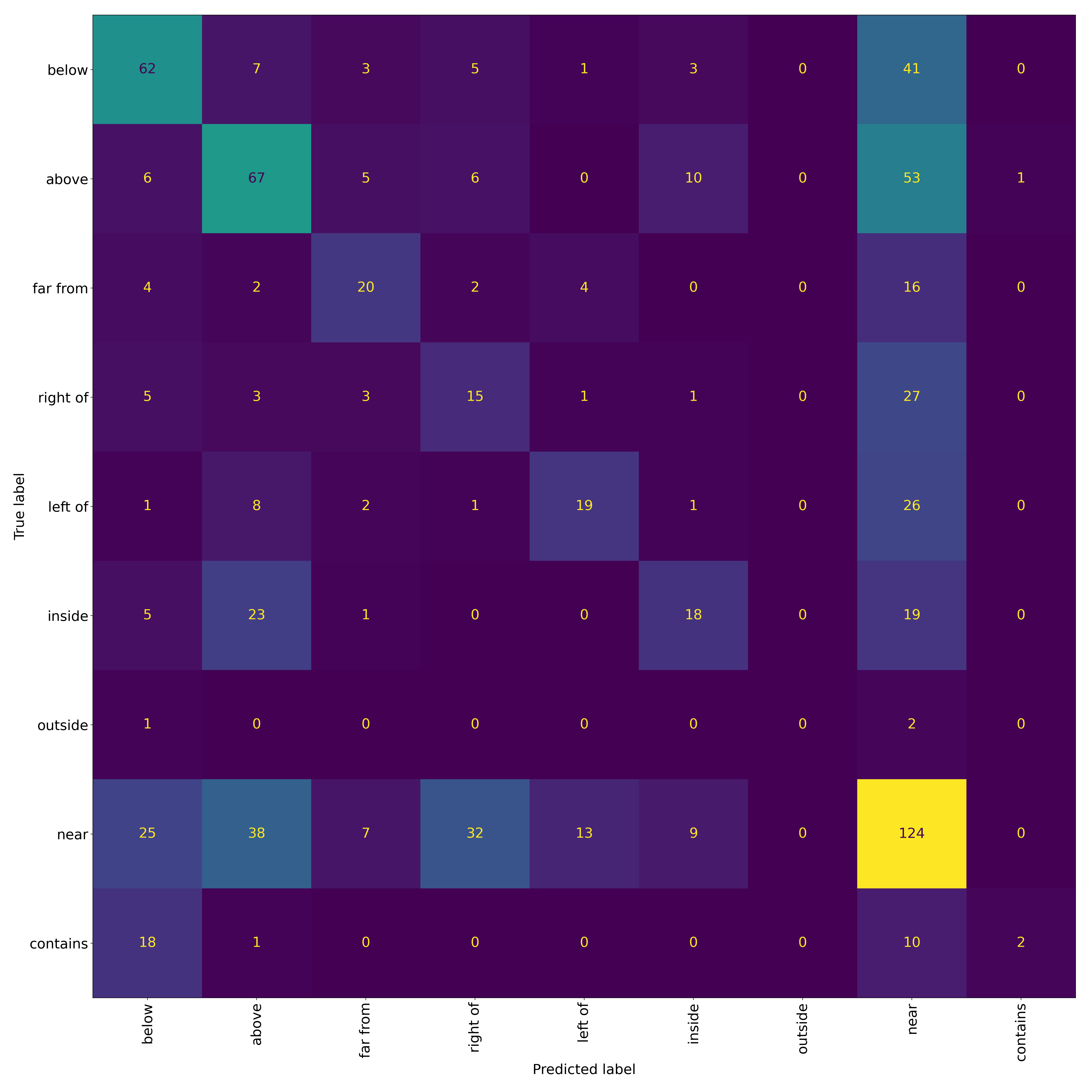}
                \caption{\textbf{After} re-ranking}
                \label{fig:subfig2}
        \end{subfigure}\hfill
        \caption{Confusion matrices for spatial relationship classification}
        \label{fig:fig2}
\end{figure*}

\subsection{GPV Localization Confidences (\texttt{Multimodal Relevance/Objectness Scores})}
Provided in Figure 7.
\label{sec:gpv}

\begin{figure*}[!h]
\centering
        \begin{subfigure}[b]{0.4\textwidth}
                \centering
                \includegraphics[scale=0.45]{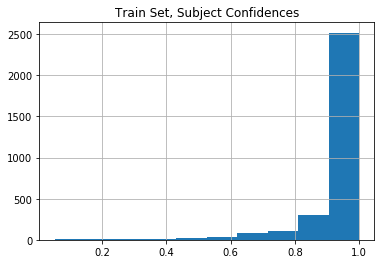}
                \caption{}
                \label{fig:subfig1}
        \end{subfigure}\hfill
        \begin{subfigure}[b]{0.4\textwidth}
                \centering
                \includegraphics[scale=0.45]{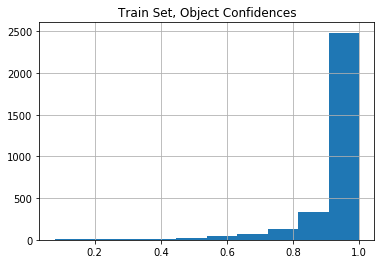}
                \caption{}
                \label{fig:subfig2}
        \end{subfigure}\hfill
        \begin{subfigure}[b]{0.4\textwidth}
                \centering
                \includegraphics[scale=0.45]{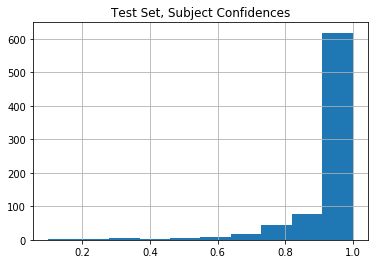}
                \caption{}
                \label{fig:subfig3}
        \end{subfigure}\hfill
        \begin{subfigure}[b]{0.4\textwidth}
                \centering
                \includegraphics[scale=0.45]{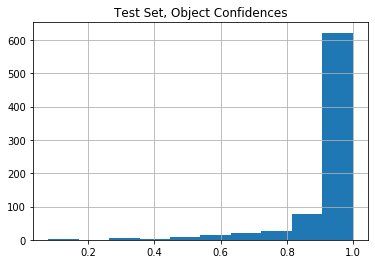}
                \caption{}
                \label{fig:subfig4}
        \end{subfigure}\hfill
        \caption{GPV Localization subject and object confidence scores for our train and test splits}
        \label{fig:fig2}
\end{figure*}

\end{document}